\documentclass[letterpaper]{article} 
\usepackage{aaai23}  
\usepackage{times}  
\usepackage{helvet}  
\usepackage{courier}  
\usepackage[hyphens]{url}  
\usepackage{graphicx} 
\usepackage{natbib}  
\usepackage{caption} 
\usepackage{algorithm}
\usepackage{algorithmic}

\usepackage{newfloat}
\usepackage{listings}
\DeclareCaptionStyle{ruled}{labelfont=normalfont,labelsep=colon,strut=off} 
\floatstyle{ruled}
\newfloat{listing}{tb}{lst}{}
\floatname{listing}{Listing}
\usepackage{pifont}
\usepackage{microtype}
\usepackage{url}

\usepackage{amsmath}
\usepackage{stmaryrd}
\usepackage{paralist}
\usepackage{subfig}
\usepackage{dashrule}
\usepackage{booktabs}
\usepackage{tabularx}
\usepackage{multirow}
\usepackage{xcolor}
\usepackage{soul}
\usepackage{amsfonts}
\usepackage{colortbl}
\usepackage{arydshln}
\usepackage{amssymb}
\usepackage{verbatim}
\usepackage{CJKutf8}

\title{Towards Reliable Neural Machine Translation with Consistency-Aware Meta-Learning}
\author{
    Rongxiang Weng\textsuperscript{\rm 1},
    Qiang Wang\textsuperscript{\rm 2,3},
    Wensen Cheng\textsuperscript{\rm 1},
    Changfeng Zhu\textsuperscript{\rm 1},
    Min Zhang\textsuperscript{\rm 1}
}
\affiliations{
    \textsuperscript{\rm 1}Soochow University, Suzhou, China
    \\
    \textsuperscript{\rm 2}Zhejiang University, Hangzhou, China \\
    \textsuperscript{\rm 3}RoyalFlush AI Research Institute, Hangzhou, China\\
    wengrongxiang@gmail.com, wangqiangneu@gmail.com, vinson7973@gmail.com, 	bajiuchangfeng@126.com, minzhang@suda.edu.cn
}

\usepackage{bibentry}

\begin{document}

\maketitle

\begin{abstract}
Neural machine translation (NMT) has achieved remarkable success in producing high-quality translations. However, current NMT systems suffer from a lack of reliability, as their outputs that are often affected by lexical or syntactic changes in inputs, resulting in large variations in quality. This limitation hinders the practicality and trustworthiness of NMT. A contributing factor to this problem is that NMT models trained with the \textit{one-to-one} paradigm struggle to handle the \textit{source diversity} phenomenon, where inputs with the same meaning can be expressed differently. In this work, we treat this problem as a bilevel optimization problem and present a consistency-aware meta-learning (CAML) framework derived from the model-agnostic meta-learning (MAML) algorithm to address it. Specifically, the NMT model with CAML (\textit{named} \textsc{CoNMT}) first learns a consistent meta representation of semantically equivalent sentences in the outer loop. Subsequently, a mapping from the meta representation to the output sentence is learned in the inner loop, allowing the NMT model to translate semantically equivalent sentences to the same target sentence. We conduct experiments on the NIST Chinese to English task, three WMT translation tasks, and the TED M2O task. The results demonstrate that \textsc{CoNMT} effectively improves overall translation quality and reliably handles diverse inputs.
\end{abstract}

\section{Introduction}
Neural machine translation (NMT) based on the \textit{encoder-decoder structure}~\cite{Bahdanau2015Neural,sutskever2014sequence} and \textit{attention mechanism}~\cite{Luong2015Effective}, has achieved prominent success in the vast majority of translation tasks~\cite{Cho2014Learning,vaswani2017attention} in recent years. Most NMT models are trained by parallel data in an end-to-end manner. In a nutshell, the encoder encodes an input sentence into sequential hidden states as contextual representation. Then, the decoder fetches the contextual representation through cross attention and generates the corresponding target sentence.

\begin{figure}[t]
    \centering
    \includegraphics[scale = 0.38]{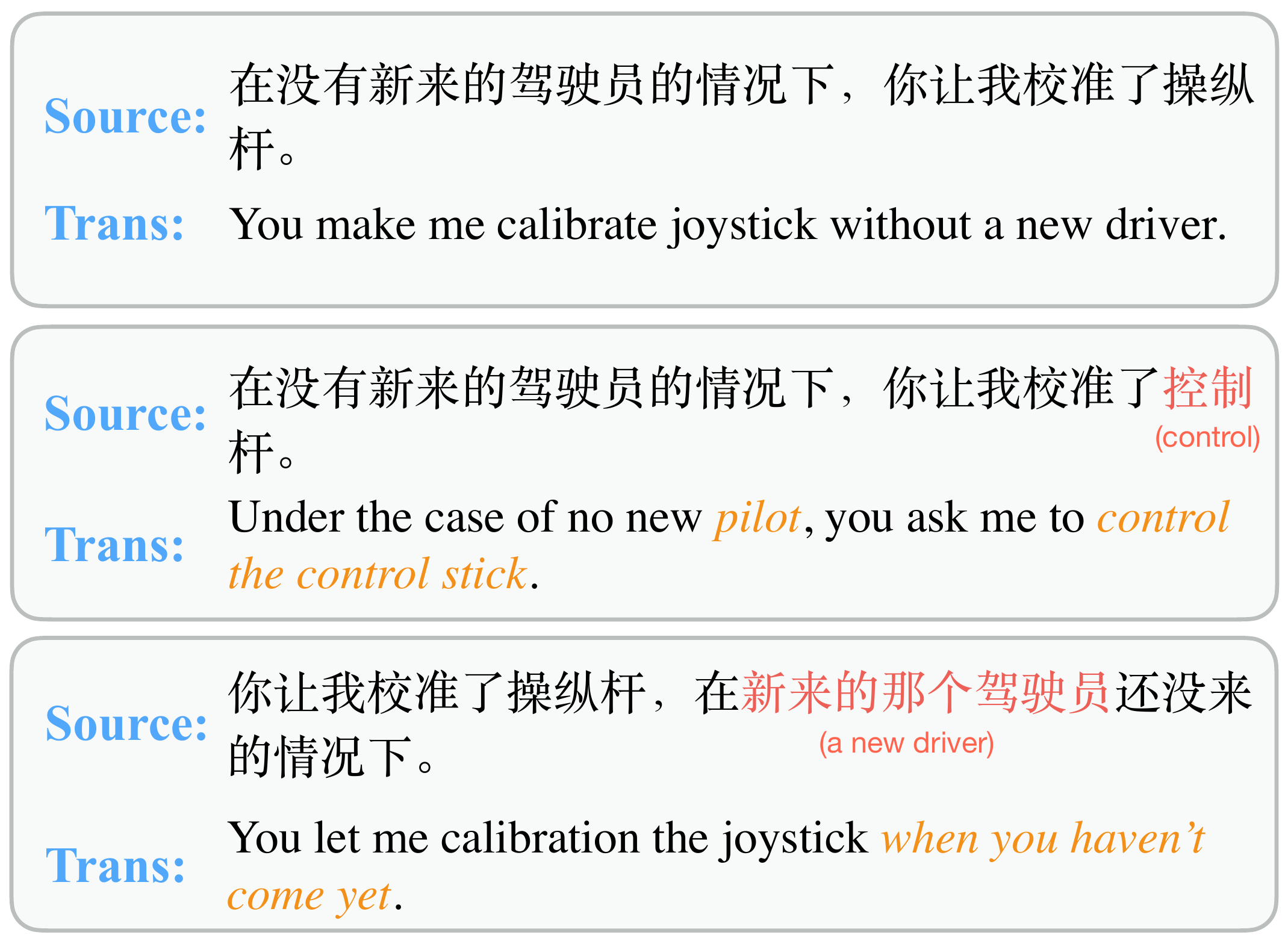}
    \caption{Translations (Trans) from semantically equivalent inputs (Source). Compared to the first one, Some \textit{semantically irrelevant} modifications in the following inputs bring apparent changes in the outputs.}
    \label{fig: transformer}
\end{figure}
 
However, many previous studies have shown that NMT models are highly susceptible to lexical or syntactic changes in inputs~\cite{ott2018analyzing,he2020structure}. This means that altering the structure of the input sentence or even just a few words, without changing the semantics, can result in a significant change in the output, frequently leading to inaccurate translations~(See Figure \ref{fig: transformer}). One reason for this is that the \textit{one-to-one} training paradigm conflicts with the \textit{many-to-many} characteristic of natural language. We define the aforementioned phenomenon as the \textit{source diversity} problem and focus on the \textit{many-to-one} situation in this work. In reality, the \textit{many-to-one} mapping is commonplace in NMT. For example, individuals may express the same concept in different ways, and even two descriptions of the same concept from a single individual may vary significantly. When NMT correctly handles the source diversity, it will become more reliable in diversified application scenarios.

Some recent works have noticed this problem and sought to address it. \citet{nguyen2019data} proposed a data augmentation method, which back-translates the original sentence by several machine translation systems, and then directly adds them into the model training process. However, it does not constrain the NMT model, leaving them to deal only with synthetic data. 
\citet{wei2020uncertainty} proposed a variational method to model uncertainty, so as to map semantically equivalent sentences into a sentence representation.
However, this method is tendentious to get a trivial solution, where all sentences are embedded in a meaningless vector.
Moreover, studies on the robustness of NMT can also mitigate this problem. 
\citet{liu2019robust} incorporated noisy input with similar pronunciations into NMT training to address \textit{homophone noise}.
\citet{cheng2019robust,cheng2020advaug} proposed to use adversarial examples to address \textit{synonym problem}. However, they only focus on the noisy input, part of the source diversity phenomenon.

In this work, we divide this source diversity problem into two subproblems:
The first one is \textit{how to learn a consistent context representation from semantically equivalent sentences}; 
The second one is \textit{how to map representation to the ground-truth outputs}. 
Following this intuition, we present a consistency-aware meta-learning (CAML) framework based on the model-agnostic meta-learning (MAML) algorithm~\cite{finn2017model} to address it. 
Unlike the traditional MAML, our CAML contains two task-specific training objectives to constrain the meta representation. Concretely, we design 1). a sentence-level objective to force semantically equivalent sentences to generate each other; 2). a word-level objective to constrain the output distributions from these sentences to be similar. CAML can promote NMT to learn a consistent meta representation from semantically equivalent samples in the outer loop. After this, NMT will focus on learning a mapping from the meta representation to the output sentence in the inner loop. 

To illustrate the effectiveness of the proposed CAML, we conduct experiments on widely used machine translation tasks with different settings. Specifically, we first evaluate our approach on the WMT14 English to German  (En$\rightarrow$De), WMT17 Chinese to English (Zh$\rightarrow$En), and WMT16 Romanian to English (Ro$\rightarrow$En) tasks, and gets more than 1 BLEU score gain in all of them. Further, we apply our approach to three varieties: non-autoregressive generation, multilingual translation, and larger model structure. The empirical results are indicative of a good versatility of our method. Finally, we make a simulation experiment, which demonstrating that our approach effectively alleviates the influence of the different forms of inputs.

\section{The Proposed Approach}
In this section, we will briefly introduce NMT based on the Transformer~\citep{vaswani2017attention} structure and the application of MAML~\cite{finn2017model} to NMT. Then, we will describe the proposed CAML based on the Transformer backbone in detail.

\subsection{Background and Notations}

\paragraph{Neural machine translation.} Given a \textit{source-target} sentence pair $\{x,y\} $ from the parallel data-set $\mathcal{B}$, where $|\mathcal{B}|$ is the number of sentence pairs. The paradigm of NMT is: 
\begin{align}
\label{eq: nmt}
P(y|x) \propto \textsc{Dec}(\textsc{Enc}(x;\theta_{e});\theta_{d}),
\end{align}
where the $\textsc{Enc}(\cdot)$ and $\textsc{Dec}(\cdot)$ are encoder and decoder networks respectively, and $\theta_{e}$ and $\theta_{d}$ are corresponding parameters. In Transformer, they are composed of multiple self-attention layers. Then the NMT model is optimized by minimizing the negative log-likelihood, denoted by:
\begin{align}
\label{eq: loss}
\mathcal{L}_{\text{N}}(y,f_{\theta}(x)) =-\mathbb{E}_{\{x,y\} \sim \mathcal{B}} \left [\log P(y|x;\theta) \right ],
\end{align}
where $\theta=\{\theta_e,\theta_d\}$, $f_{\theta}(x)$ denotes the model prediction of $x$ with model parameters $\theta$.

\paragraph{MAML for NMT.} 

Model-agnostic meta-learning (MAML) is an widely used meta-learning (or \textit{learning to learn}) algorithm, which divides the training process into two stages~\cite{finn2017model}, named \textit{meta-train} and \textit{meta-test}. In \textit{meta-train} step, a general model is learned from different data (called supported set), while a task-specific data (called target set) is used in the \textit{meta-test} step. 

In our setup, we take the original parallel data $\{x,y\}$ as the target set, and construct the support set by introducing a set $\textbf{x}^{s}$ which is semantically equivalent to $x$. $\textbf{x}^{s}=\{x^{s}_1,\cdots,x^{s}_i,\cdots,x^{s}_I\}$, where $I$ is the number of sentences. The $\textbf{x}^{s}$ is generated from a sampling function $\varphi(x,y)$, which will be described in the Section \ref{sec: data}. Following \citet{gu2018meta,li2018learning}, we can readily employ MAML on NMT to utilize $\textbf{x}^{s}$.
In this way, we first update the current model parameters from $\theta$ to $\theta'$ by the gradients in the meta-train step:
\begin{align}
    \theta' \leftarrow \theta - \alpha \nabla_{\theta}&\left \{\mathcal{L}_{\text{Train}}(y_{i},f_{\theta}(x_{i}^s))  \right \},
\end{align}
where $\alpha$ is the learning rate and $\mathcal{L}_{\text{Train}}$ is the loss function for the meta-train step. We can calculate $\mathcal{L}_{\text{Train}}$ through the support set as:
\begin{align}
    \label{eq: meta-train-t}
    \mathcal{L}_{\text{Train}}(y_{i},f_{\theta}(x^{s}_{i}))=\frac{1}{I}\sum_{i}^{I}\left \{\mathcal{L}_{\text{N}}(y_{i},f_{\theta}(x_{i}^s))\right \}.     
\end{align}
Given $\theta'$, we further leverage the target set to update the model parameters through the meta-test step:
\begin{align}
    \label{eq: meta-test-t}
    \mathcal{L}_{\text{Test}}(y,&f_{\theta}(x))=\mathcal{L}_{\text{N}}(y,f_{\theta'}(x)).
\end{align}

\subsection{Consistency-aware Meta-learning for NMT}
The ideal case for increasing the reliability of NMT is that semantically equivalent sentences have the same representation in the semantic space, which allows the model to generate stable outputs from various inputs. 
Following this motivation, the proposed CAML concentrates on promoting NMT to learn a consistent meta representation from semantically equivalent samples and to translate from the meta representation to the ground truth. 
In this section, we will first introduce two simple training objectives in the CAML, which constrain the contextual representation in different aspects. 
Then, we will expound on how to incorporate the CAML into the training process to build a reliable NMT model. 

\paragraph{Modeling consistency through multi-level constraints.}
In Transformer, the output is decided by the contextual representation encoded from the encoder. Namely, when the representations from sentences in $\textbf{x}^{s}$ are equivalent: 
\begin{align}
||\textsc{Enc}(x^{s}_{i};\theta_{e})-\textsc{Enc}(x^{s}_{j};\theta_{e})||_{2}^{2}=0,  \{x^{s}_{i},x^{s}_{j}\} \subset \textbf{x}^s,
\end{align}
where the NMT model can generate the same output by the contextual representation.  
However, it is hard to optimize this objective due to the variable sentence structure.
From another perspective, we consider two training objectives of different granularity to optimize the NMT model.

The first is the consistency of the sentence representation, which we consider to be optimized by a partial reconstruction manner.
We randomly sample a sentence pair $\{x_i^s,x_j^s\}$ from $\textbf{x}^{s}$ and mask the same part of them, which are denoted as $\{x_i^m,x_j^m\}$.
Then, the NMT model is asked to generate one when given another. 
Specifically, the sentence level training loss is formalized as: 
\begin{align} 
    \mathcal{L}&_{\text{S}}(x^{s}_{j},x^{s}_{i})= \mathcal{L}_{\text{N}}(x^{s}_{j},f_{\theta}(x^{s}_{i}))+\mathcal{L}_{\text{N}}(x^{s}_{i},f_{\theta}(x^{s}_{j}))= \\
    &-\mathbb{E}_{\{x_{i}^{s},x_{j}^{s}\} \sim \varphi(\cdot)}\mathbb{E}_{\{x,y\} \sim \mathcal{B}}  [\log P(x_{j}^{m}|x_{i}^{s};\theta)P(x_{i}^{m}|x_{j}^{s};\theta)]. \nonumber
\end{align}
Here, $\mathcal{L}_{\text{S}}$ encourages the contextual representation to recover sentences that have similar semantics as itself. 
This is almost equal to enabling NMT to generate the same sentence representations between semantically equivalent sentences.

\begin{algorithm}[t]
    \caption{The training process of NMT with CAML}
    \label{alg: training}
    \textbf{Input}: Parallel data set $\mathcal{B}$; Sampling function: $\varphi(\cdot)$; Epoch $E$  \\
    \textbf{Output}: The NMT model $f_{\theta}(\cdot)$ \\
    \vspace{-0.4cm}
    \begin{algorithmic}[1] 
    \FOR{$e = 0$ to $E$}
    \FOR {$\{x,y\}$ in $\mathcal{B}$}
    \STATE Optimizing $f_{\hat{\theta}}(\cdot)$ by the Equation \ref{eq: loss} \\
    \ENDFOR \\
    \ENDFOR
    \FOR{$e = 0$ to $E$}
    \FOR {$\{x,y\}$ in $\mathcal{B}$}
    \STATE Sampling $\textbf{x}^{s}$ by the $\varphi(x,y)$ \\
    \FOR {$\_$ in T}
    \STATE Optimizing $f_{\theta'}(\cdot)$ by the Equation \ref{eq: meta-train} \\
    \ENDFOR \\
    \STATE Optimizing $f_{\theta}(\cdot)$ by the Equation \ref{eq: meta-test}
    \ENDFOR 
    \ENDFOR
    \STATE \textbf{return} $f_{\theta}(\cdot)$
    \end{algorithmic}
\end{algorithm}

Besides modeling the sentence-level semantics embedded in the sequential representations, we need to treat the difference in syntactic structure, determining how the decoder gets the correct contextual information. 
In contrast, instead of exploiting the syntactic information, which needs human-annotated data, we adopt a word-level training objective as an alternative to making the output distribution generated by one sentence governed by the others.


Specifically, we constrain the output distributions from semantically equivalent sentences to be similar at each decoding step.
Formally, the word-level loss function is summarized as:
\begin{align}
    \mathcal{L}_{\text{W}}&(f_{\tilde{\theta}}(x),f_{\theta}(x^{s}_{i}))=  \\ &- \mathbb{E}_{x_{i}^{s} \sim \varphi(\cdot)}\mathbb{E}_{\{x,y\} \sim \mathcal{B}} [P(y|x;\tilde{\theta})  \log P(y|x_{i}^{s};\theta)], \nonumber
\end{align}
where $\tilde{\theta}$ is the copy of the NMT parameters, which will not be updated by the gradient. For different forms of input, the NMT decoder trained by this objective can get the similar contextual information at each step, avoiding the the influence of the structural differences.


\paragraph{Meta-learning with $\mathcal{L}_{\text{S}}$ and $\mathcal{L}_{\text{W}}$.}
Following the MAML framework~\cite{finn2017model}, we sample $T$ pairs from $\textbf{x}^{s}$ as the \textit{support set}, and then use the $\{x^{s}_{i},x,y\}$ as the \textit{target set}. The training function of the meta-train step can be summarized as follows: 
\begin{align}
    \label{eq: meta-train}
    \mathcal{L}_{\text{Train}}&(x^{s}_{\setminus i},f_{\theta}(x^{s}_{i}))=\\ 
    &\beta  \sum^{T}\left \{\mathcal{L}_{\text{S}}(x^{s}_{\setminus i},f_{\theta}(x^{s}_{i}))+\mathcal{L}_{\text{W}}(f_{\theta}(x^{s}_{\setminus i}),f_{\theta}(x^{s}_{i}))\right \},   \nonumber    
\end{align}
where $x^{s}_{\setminus i}$ is a sentence from $\textbf{x}^{s}$ which not equal to $x^{s}_i$, $\beta$ is the learning rate for the meta-train step.
Compared to the Eq. \ref{eq: meta-train-t}, we use the aforementioned self-supervised training objectives instead of the NMT training loss (\textit{i.e.}, $\mathcal{L}_{\text{T}}$) in this step. Our goal is to facilitate the model to pay attention to learning the contextual representation in the outer loop rather than directly learning the translation from noisy input data.

\begin{figure}[t]
    \centering
    \includegraphics[scale = 0.80]{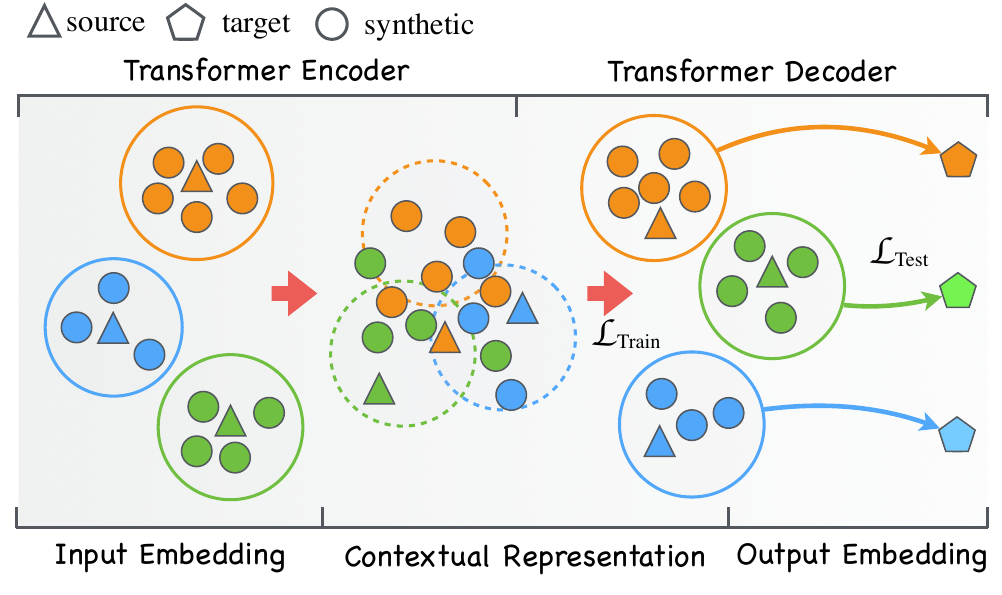}
    \caption{\label{fig: model} An intuitive illustration of the impact of the proposed CAML on the contextual representation of NMT.}
\end{figure}

The next step of the CAML is to learn to translate based on contextual representation. An intuitive approach is to adopt Eq. \ref{eq: meta-train} to train the model. However, it may cause the \textit{forgetting} problem since the objective functions of the two steps are different. To avoid this problem, we integrate $\mathcal{L}_{\text{S}}$ and $\mathcal{L}_{\text{W}}$ into the meta-test step as a regularization term to maintain the representation consistently. 
Specifically, the meta-test step in the CAML is reorganized as: 
\begin{align}
    \label{eq: meta-test}
    \mathcal{L}_{\text{Test}}&(y,f_{\theta}(x))=\gamma \mathcal{L}_{\text{N}}(y,f_{\theta}(x)) \\
    &+\underbrace{\epsilon\left \{ \mathcal{L}_{\text{S}}(x,f_{\theta{'}}(x^{s}_{i}))+\mathcal{L}_{\text{W}}(f_{\theta}(x),f_{\theta{'}}(x^{s}_{i})) \right \}}_{\text{maintaining consistency}}.   \nonumber    
\end{align}
The coefficients $\gamma$ and $\epsilon$ are used to balance the preference among these losses, which are empirically set to 1 and 0.5, respectively. Here, the updated parameters $\theta'$ are computed by $T$ gradient descent updates on the support task, which equals $2I$. The one gradient update in the meta-train step is:
\begin{align}
    \theta' \leftarrow \theta - \gamma \nabla_{\theta}&\left \{\mathcal{L}_{\text{S}}(x_{\setminus i},f_{\theta}(x_{i}))) +\mathcal{L}_{\text{W}}(f_{\theta}(x_{\setminus i}),f_{\theta}(x_{i})) \right \},\nonumber 
\end{align}
where $\gamma$ is the learning rate of the meta-train step. The overall training algorithm of our approach is shown in Algorithm \ref{alg: training}. An intuitive illustration is shown in Figure \ref{fig: model}.

\subsection{Semantically Equivalent Sentences}
\label{sec: data}
A key factor of our approach is how to build the $\textbf{x}^{s}$ by the $\varphi(\cdot)$. 
It is difficult to obtain perfect semantically equivalent sentences by data augmentation methods.
Here, we use two steps to construct semantically similar sentences as approximations. Coincidentally, previous studies show that noisy samples enhance the performance of meta-learning~\cite{wang2020training,yao2021meta}. It is worthy to note that our work focuses on the training strategy of NMT, so we only use two common data augmentation approaches. Any advanced approaches can be easily applied here~\cite{wang2018switchout,hoang2018iterative,nguyen2019data}.

\paragraph{Step 1: replacing words with alignment relationships.} Given the sentence $x$, we sample and replace 20\% words with similar words. More specifically, for a selected word in $x$, we choose 100 words from the vocabulary $\mathcal {V} $ ordered by the frequency of a word alignment model, which is trained on the same parallel data.\footnote{https://github.com/clab/fast$\_$align} Then, we form these words as a subset and randomly choose one to replace the original word.

\paragraph{Step 2: noisy round-trip translation.}
Then, we use noisy round-trip translation to process the data from Step 1 further. Given sentences from Step 1, we add random vectors sampled from the standard normal distribution to them and translate them to target language sentences $\textbf{y}'=\{y'_1,\cdot,y'_j,
\cdot,y'_J\}$, where $J$ is the number of sentences. 
We select several sentences from the $\textbf{y}'$ according to the weight $\textsc{EditDist}(y'_j,y)*\textsc{Overlap}(y'_j,y)$. The $\textsc{EditDist}(\cdot)$ is edit distance and $\textsc{Overlap}(\cdot)$ is word overlap ratio. Then, these sentences are back-translated to the source language to compose the $\textbf{x}^{s}$.

\begin{table*}[t]
    \centering
    \begin{tabular}{clccccccc}
    \toprule
    \#&\textbf{Model}&\textbf{MT06}&\textbf{MT02}&\textbf{MT03}&\textbf{MT04}&\textbf{MT05}&\textbf{MT08}&\textbf{Avg} \\
    \midrule
    1&\citet{vaswani2017attention}&44.57&45.49&44.55&46.20&44.96&35.11&43.26 \\
    2&\citet{ebrahimi2018adversarial}&45.28&45.95&44.68&45.99&45.32&35.84&43.56 \\
    3&\citet{cheng2019robust}& 46.95&47.06&46.48&47.39&46.58&37.38&44.98\\
    4&\citet{hao2019multi}&44.00&N/A&43.98&45.60&44.28&N/A&N/A\\
    5&\citet{cheng2020advaug}& 49.26&49.03&47.96&48.86&49.88&39.63&47.07\\
    \midrule
    6&\citet{sennrich2016improving}$^*$&46.20&47.78&46.93&47.80&46.81&36.79&45.22\\
    7&\citet{cheng2019robust}$^*$&47.74&48.13&47.83&49.13&49.04&38.61&46.55\\
    8&\citet{wei2020uncertainty}$^*$ &48.88&N/A&N/A&49.15&49.21&\textbf{40.94}&N/A\\
    9&\citet{cheng2020advaug}$^*$&\textbf{49.98}&50.34&49.81&50.61&50.72&40.45&48.39\\
    \midrule
    10&Transformer&45.24&46.89&46.32&46.83&45.72&36.31&44.41 \\
    11&Transformer$_{\text{sync}}^*$&46.57&47.13&47.69&47.51&46.88&39.67&45.77 \\
    12&\textsc{CoNMT}&49.41&49.83&48.83&49.51&50.03&39.99&47.64 \\
    13&\textsc{CoNMT}$_{\text{sync}}^*$&49.74&\textbf{50.79}&\textbf{50.53}&\textbf{50.96}&\textbf{51.27}&40.88&\textbf{48.89} \\
    \bottomrule
    \end{tabular}
    \caption{The effectiveness of our approach (\textsc{CoNMT}) on the NIST Zh$\rightarrow$En task. ``$*$'' indicates the model uses extra monolingual data-set. ``Avg'' is the average score of all test sets.}
    \label{tab: mt_zhen}
 \end{table*} 
 
\section{Experiments}
\subsection{Implementation Detail}
\paragraph{Data-set.} We first conduct experiments on the NIST Chinese to English (Zh$\rightarrow$En) task and three widely-used WMT translation tasks: WMT14 English to German (En$\rightarrow$De), WMT16 Romanian to English (Ro$\rightarrow$En), and WMT18 Chinese to English (Zh$\rightarrow$En). On the NIST Zh$\rightarrow$En task, we use NIST 2006 (\texttt{MT06}) as the dev set and NIST 2002 (\texttt{MT02}), 2003 (\texttt{MT03}), 2004 (\texttt{MT04}), 2005 (\texttt{MT05}), 2008 (\texttt{MT08}) as the test sets. 
On the En$\rightarrow$De task, we use {\texttt{newstest2013}} as the dev set and \texttt{newstest2014} as the test set. On the Ro$\rightarrow$En task, we use {\texttt{newstest2015}} as the dev set and \texttt{newstest2016} as the test set. 
On the Zh$\rightarrow$Ee task, we use {\texttt{newsdev2017}} as the dev set, and \texttt{newstest2017} as the test set. 
Moreover, we evaluate our method on the TED multilingual machine translation task. The overall data-set is divided into two training sets according to language diversity, which is named \textit{related} and \textit{diverse}. The detail is shown in \citet{wang2020balancing}. We only conduct the many-to-one (M2O), which translates eight languages to English.

\paragraph{Setting.} 
We adopt the Transformer-\textit{base} for the NIST Zh$\rightarrow$En and WMT Ro$\rightarrow$En, Transformer-\textit{big} for the WMT Zh$\rightarrow$En, and both \textit{base} and \textit{big} for the WMT En$\rightarrow$De. 
We apply the byte pair encoding (BPE)~\cite{sennrich2015neural} to all language pairs and limit vocabulary size to 32K. All out-of-vocabulary words were mapped to the \textit{UNK}. In order to avoid the problem of vocabulary mismatch, we adopt an additional four-layer parallel decoder~\cite{ghazvininejad2019mask} for the $\mathcal{L}_\text{S}$.
We set label smoothing as 0.1 and dropout rate as 0.1. The Adam is adopted as the optimizer, and the $\beta1/\beta2$ is set as $0.9/0.98$ for the \textit{base} setting and $0.9/0.998$ for the \textit{big} setting.
The initial learning rate is 0.001. We adopt the warm-up strategy with 4000 steps. The learning rate $\gamma$ of \textit{meta-train} remains one-tenth of NMT. Other settings are followed as \citet{vaswani2017attention}.

We implement our approach on \textit{fairseq}\footnote{https://github.com/pytorch/fairseq}.
We use 4 V100 GPUs and accumulate the gradient four iterations on the WMT En$\rightarrow$De and Zh$\rightarrow$En. Other tasks run on 2 V100 GPUs.
We use beam search as the decoding algorithm and set the beam size as 5 and the length penalty as 0.6.
For a fair comparison, we calculate the \textit{case-sensitive tokenized} BLEU with the \textit{multi-bleu.perl} script for the NIST Zh$\rightarrow$En, and use the \textit{sacreBLEU}\footnote{https://github.com/mjpost/sacreBLEU}\footnote{BLEU+case.mixed+lang.\$\{Src\}-\$\{Trg\}+num-
refs.1+smooth.exp+test.\$\{Task\}+tok.13a+version.1.5.1} to calculate \textit{case-sensitive} BLEU~\cite{Papineni2002bleu} for all WMT and TED tasks.

\subsection{Results on NIST Zh$\rightarrow$En Task}
In order to compare with the related work, we first carried out experiments on the NIST Zh$\rightarrow$En task with the same setting~\cite{wei2020uncertainty,cheng2020advaug}. 
Following \citet{cheng2020advaug}, we sample 1.25M English sentences from the Xinhua portion of the gigaword corpus as the extra corpus to enhance the model.\footnote{https://catalog.ldc.upenn.edu/LDC2003T05} 

The overall results are shown in Table \ref{tab: mt_zhen}.
We divide the experiment into three parts. Firstly, we compare our model to the Transformer baselines. We implement the vanilla Transformer-base trained by the parallel data (Transformer, Row 10) and both the parallel and synthetic data (Transformer$_{\text{sync}}$, Row 11), respectively. The synthetic data brings a 1.36 BLEU score gain.
Transformer with the proposed CAML (\textsc{CoNMT}, Row 12) attains 47.64 BLEU. When incorporating the synthetic data, our approach improves 1.24 to 48.89 BLEU (\textsc{CoNMT}$_{\text{sync}}$, Row 13). Summarily, our approach gets 3.23/3.12 absolute improvements compared to the corresponding baselines on the NIST Zh$\rightarrow$En task.

Then, we compare our approach to several related work including the robust NMT~\cite{ebrahimi2018adversarial,cheng2019robust,cheng2020advaug}, data-augmentation~\cite{wei2020uncertainty}, and enhancing the contextual representation~\cite{hao2019multi}. In the same way, we also report the results from the model trained with or without synthetic data.
Our approach achieves the best performance on both settings. Furthermore, our method outperforms the AdvAug by 0.57/0.50 BLEU and achieves state-of-the-art in four test sets.

\begin{table}[t]
    \centering
    \begin{tabular}{clcc}
    \toprule
    \#&\textbf{Model}&\textbf{Trans. Base}&\textbf{Trans. Big}\\
    \midrule
    1&\citet{vaswani2017attention} &27.30&28.40 \\
    2&\citet{li2018multi}&28.51&29.28\\
    3&\citet{cheng2018towards}$^*$&28.09&N/A\\
    4&\citet{gao-etal-2019-soft}$^*$&N/A&29.70\\
    5&\citet{jiao2020data}$^*$&28.30&29.20 \\
    \midrule
    6&\text{Transformer} &27.53&28.73 \\
    7&\text{Transformer}$_{\text{sync}}^*$ &28.82&29.71 \\
    8&\textsc{CoNMT} &28.64$^\ddagger$&29.58$^\dagger$ \\
    9&\textsc{CoNMT}$_{\text{sync}}^*$ &\textbf{30.22}$^\ddagger$&\textbf{30.87}$^\ddagger$ \\
    \bottomrule
    \end{tabular}
    \caption{The effectiveness of our approach (\textsc{CoNMT}) on the En$\rightarrow$De task. ``$\dagger$/$\ddagger$'' indicate the model is significantly better than the baseline ($p<0.05/0.01$)~\cite{DBLP:journals/corr/abs-1903-07926}. ``$*$'' indicates using extra monolingual data. ``Trans. Base'' and ``Trans. Big'' are the \textit{base} and \textit{big} setting, respectively.}
    \label{tab: mt_ende}
 \end{table}

 \begin{table}[t]
      \centering
      
      \begin{tabular}{clcc}
      \toprule
      \#&\textbf{Model}& \textbf{Zh}$\rightarrow$\textbf{En}&\textbf{Ro}$\rightarrow$\textbf{En}\\
      
      \midrule
      1&\citet{vaswani2017attention}&25.21& 31.90\\
      2&\citet{wei2020uncertainty} &26.48 & N/A \\
      3&\citet{yang2019context}&25.15 &N/A \\
      4&\citet{zhou-etal-2020-uncertainty}&25.04&N/A \\
      5&\citet{gu2018meta}&N/A &31.76\\
      \midrule
      6&\text{Transformer} &25.37&31.93 \\
      7&\text{Transformer}$_{\text{sync}}^*$ &26.63&36.94 \\
      8&\textsc{CoNMT}  &26.38&33.72\\
      9&\textsc{CoNMT}$_{\text{sync}}^*$  &\textbf{27.49}&\textbf{38.44}\\
        \bottomrule
      \end{tabular}
      \caption{The effectiveness of our approach (\textsc{CoNMT}) on the WMT Zh$\rightarrow$En and WMT16 Ro$\rightarrow$En. ``$*$'' indicates that the model uses extra monolingual data. We use the Transformer-big on the Zh$\rightarrow$En and Transformer-base on  the Ro$\rightarrow$En.
      }
      \label{tab: mt_zhro}
  \end{table}

\subsection{Results on WMT Tasks\cite{aguilar2020knowledge} }
We further evaluate our approach on three widely-used WMT benchmarks, including WMT14 En$\rightarrow$De, WMT17 Zh$\rightarrow$En, and WMT16 En$\rightarrow$Ro. Meanwhile, we report several previous work as a comparison.\footnote{We consider the work which measure the results by the \textit{scareBLEU} toolkit.}
The results on WMT14 En$\rightarrow$De are summarized in the Table \ref{tab: mt_ende}. 
Our model gets 28.64/29.58 BLEU in the \textit{base}/\textit{big} setting (Row 8). 
Compared to the baseline (Row 7), our model achieves 1.11/0.85 gains. 
We use 8M German monolingual sentences sampled from \citet{caswell2019tagged} to enhance the model. When combined with the synthetic data, our model gets 30.22/30.87 BLEU (Row 9), which gets 1.4/1.06 improvements and outperforms all previous work.

The results on the WMT17 Zh$\rightarrow$En and WMT16 Ro$\rightarrow$En are shown in Table \ref{tab: mt_zhro}. We use 10M monolingual sentences sampled from WMT News Craw and 2M monolingual sentences from \citet{sennrich2016improving} for the Zh$\rightarrow$En and Ro$\rightarrow$En, respectively. 
Compared to baselines, our work gets 1.01/0.86 gains on the Zh$\rightarrow$En, and 1.79/1.50 gains on the Ro$\rightarrow$En. The extensive experiments on the WMT benchmark demonstrate more powerfully the effect of the proposed CAML algorithm.

\begin{table}[t]
  \centering
  \begin{tabular}{lcc}
  \toprule
  \textbf{Model} &\textbf{Speedup} &\textbf{En}$\rightarrow$\textbf{De}  \\
  \midrule
  \citet{vaswani2017attention} &1.0x&27.30\\
  \citet{gu2017non} &2.36x &19.17 \\
  \citet{ghazvininejad2019mask} &1.7x& 27.03\\ 
  \midrule
  Transformer &1.0x&27.53 \\
  Mask-Predict &1.63x& 27.14\\ 
  \textsc{CoNMT} &1.63x&28.21 \\
  \bottomrule
  \end{tabular} 
  \caption{The results of the proposed CAML (\textsc{CoNMT}) on the NAT structure.}
  \label{tab: nat}
\end{table}

\subsection{Effectiveness on Different Varieties}
The proposed CAML framework only changes the training schema of NMT, which means it can be employed on any Transformer-based NMT tasks. 
In this section, we conduct condensed experiments on three popular NMT directions: non-autoregressive translation, multi-lingual translation, and larger model size.

\paragraph{Non-autoregressive NMT.}
Our approach is tangential to the decoding mode. Thus, whether our approach works on NAT~\cite{gu2017non,ghazvininejad2019mask} is valuable to investigate. 
Here, we employ the CAML based on the Mask-Predict~\cite{ghazvininejad2019mask} framework\footnote{The detail of this model is shown in \citet{ghazvininejad2019mask}.} and evaluate on the En$\rightarrow$De task. The results are shown in Table \ref{tab: nat}. Our work largely outperforms the Mask-Predict baseline (+1.07 BLEU) without dropping the decoding speed. 

\begin{table}[t]
  \centering
  \begin{tabular}{lcc}
  \toprule
  \textbf{Model}& \textbf{Related}&\textbf{Diverse}\\
  
  \midrule
  Uni. ($\tau=\infty$)&22.63&24.81\\
  Temp. ($\tau=5$) &24.00&26.01 \\
  Prop. ($\tau=1$) &24.88&26.68 \\
  \midrule
  \citet{wang2020balancing} &25.26&27.00 \\
  \citet{wu2021uncertainty} &26.28&27.82 \\
  \midrule
  \textsc{CoNMT}  &26.77&28.09\\
    \bottomrule
  \end{tabular}
  \caption{The results of our method (\textsc{CoNMT}) on the M2O task. \textit{Uni.}: sampling data with equal frequency. \textit{Temp.}: sampling data according to
  size exponentiated by $\tau$. \textit{Prop.}: sampling data in portions equivalent to the size of each language.}
  \label{tab: mt_m2o}
\end{table} 

\paragraph{Multilingual NMT.}
We believe that sentences from different languages with the same semantics can be regarded as semantically equivalent sentences.
To verify that, we employ our approach on the many-to-one (M2O) translation task following the setting from \citet{wang2020balancing}. We sample $I$ sentences
and use a one-to-many NMT model to translate to other languages. Then these synthetic sentences are used as training data for the meta-train step.

The results are shown in Table \ref{tab: mt_m2o}. Compared to \citet{wang2020balancing}, our approach obtains 1.51/1.09 improvements in the related/diverse language pairs. Furthermore, our approach gets better results than \citet{wu2021uncertainty}, especially in related language pairs. Empirically, our approach works better for similar language pairs than diverse ones, which indicate that it is difficult for NMT to learn consistent representations for diverse languages.

\begin{table}[t]
  \centering
  \begin{tabular}{lcc}
  \toprule
  \textbf{Model} &\#\textbf{Param} &\textbf{En}$\rightarrow$\textbf{De}  \\
  \midrule
  \citet{vaswani2017attention} &213M&28.40\\
  \citet{wei2022learning}&265M&29.80\\
  \citet{wang2019learning}&137M&29.30 \\
  \citet{zhu2020incorporating} & 500M & 30.75 \\
  \citet{raffel2020exploring} & 700M & 30.90 \\
  \midrule
  Deep Transformer &159M&29.82 \\
  mBART Fine-tuning &610M& 28.82 \\
  \textsc{CoNMT}$_\text{sync}$ &207M& 30.87 \\
  mBART \textit{w/} CAML &610M& 30.23\\
  Deep Transformer \textit{w/} CAML &159M&31.26 \\
  \bottomrule
  \end{tabular} 
  \caption{The results of the proposed method with larger model size on the En$\rightarrow$De task.}
  \label{tab: semi}
\end{table}

\paragraph{Larger model size.} 
A spontaneous confusion is whether a larger model can eliminates the source diversity problem. To verify the availability in this situation, we employ the CAML on the pre-trained mBART~\cite{liu2020multilingual} and Deep Transformer~\cite{wang2019learning} on the En$\rightarrow$De task. 
The results are shown in Table \ref{tab: semi}. Compared to the fine-tuning, our approach gets a 1.41 improvement. Then, when combined with Deep Transformer, our approach gets a 1.44 gain, outperforming all previous work. The evaluation results demonstrate that a larger model can not escape the source diversity problem, and our CAML still works in this situation.

\begin{figure}[t]
    \centering
    \includegraphics[scale = 0.20]{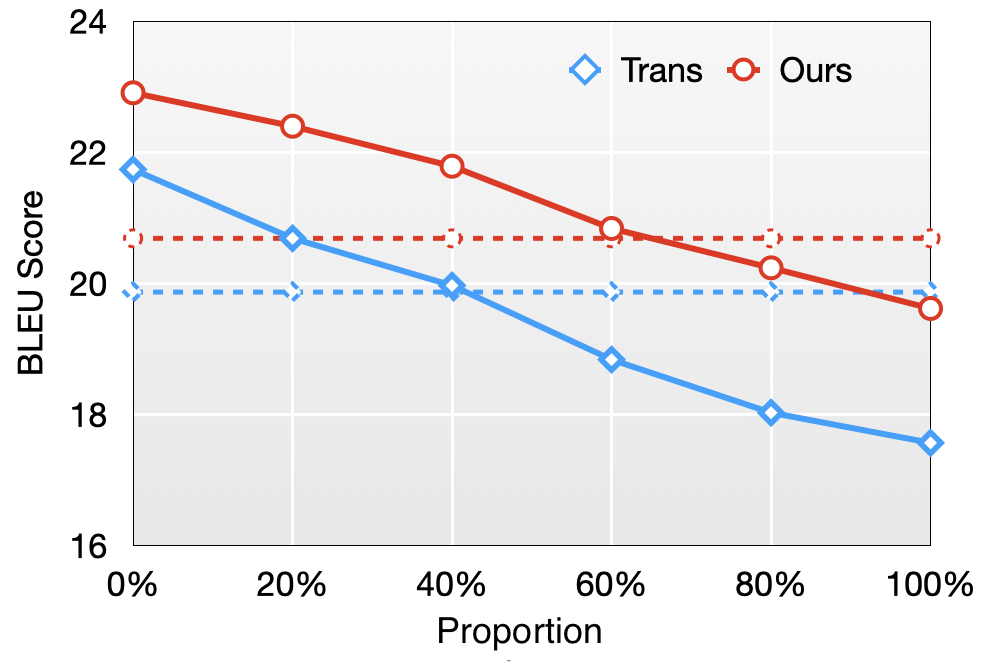}
    \caption{\label{fig: mi} The BLEU curve under the different ratio of diverse input. The dotted line is the BLEU from the original inputs.}
\end{figure}
\begin{figure}[t]
    \centering
    \includegraphics[scale = 0.28]{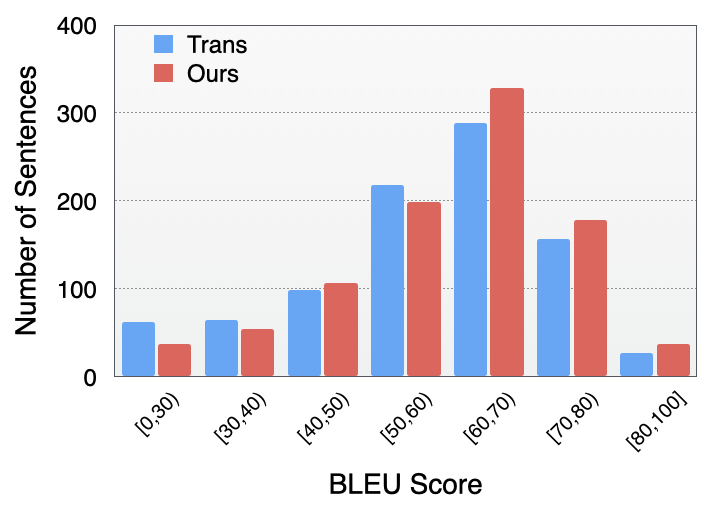}
    \caption{\label{fig: res_dis} The number of sentences divided according to the sentence BLEU.}
\end{figure}
\subsection{Analysis}
\paragraph{The influence of the diverse input.} 
We sample 300 samples from MT02-05, each one has one source sentence and four references, to appraise the ability of our approach to handle the source diversity problem. 
We ask two professional translators to translate source sentences according to the first reference and the remaining three, respectively. 
Then, we use the first reference and the corresponding translation as the standard test set, and replace input sentences from another translator in different proportions. The results are presented in Figure \ref{fig: mi}.
As the replacement proportion increases, the performance of all models decreases.
However, our model is more reliable than the baseline, which performs similarly to the original input even with a large replacement proportion.

\paragraph{Distribution across translation quality.} Then, we sample one thousand translations on the En$\rightarrow$De task and calculate the sentence-BLEU of each one.\footnote{We use the script from THUMT: https://github.com/THUNLP-MT/THUMT/blob/master/thumt/utils/bleu.py}  We divide these sentences into several groups according to the BLEU score and count the number of sentences at each group. The results are shown in Figure \ref{fig: res_dis}. 
Compared to the baseline, the translations of our model are more concentrated into interval [60-80) and have less number in interval [0-40).
The evaluation results show that the proposed CAML can accelerate the model to generate more stable results, which effectively improves the reliability of NMT.

\begin{table}[t]
  \centering
  \begin{tabular}{lcc}
  \toprule
  \textbf{Model}& \textbf{En}$\rightarrow$\textbf{De}&$\Delta$\\
  \midrule
  Transformer~\cite{vaswani2017attention} &  27.53& $-$\\
  \midrule
   \ \ $+$ Data Augmentation&27.87& +0.34\\
   \ \ $+$ MAML (Eq.\ref{eq: meta-test-t}-\ref{eq: meta-train-t})&27.41&-0.12\\
   \ \ $+$ MTL with $\mathcal{L}_{\text{W}}$ and $\mathcal{L}_{\text{S}}$ &28.01&+0.48\\
   \midrule
   
   \ \ $+$ CAML (\textsc{CoNMT})  &28.64&+1.11\\
  \ \ \ \ $+\mathcal{L}_{\text{T}}$ in Eq. \ref{eq: meta-train}&28.37&+0.84\\
  \ \ \ \ $-\mathcal{L}_{\text{S}}$ in Eq. \ref{eq: meta-train} and Eq. \ref{eq: meta-test} &28.27&+0.74\\
  \ \ \ \ $-\mathcal{L}_{\text{W}}$ in Eq. \ref{eq: meta-train} and Eq. \ref{eq: meta-test} &28.12&+0.59\\
  \bottomrule
  \end{tabular}
  \caption{Ablation study on the WMT14 En$\rightarrow$De task.}
  
  \label{tab: ablation}
\end{table}

\paragraph{Ablation study.}
To further show the effect of each module in our approach, we make an ablation study which is shown in Table \ref{tab: ablation}. 
On the one hand, we implement three methods to analyze effect of $\textbf{x}^s$, $\mathcal{L}_{\text{S}}$ and $\mathcal{L}_{\text{W}}$. 
The first one is to use $\{\textbf{x}^s,y\}$ with the original training set to train the model. 
The second one is to apply MAML as described in Eq.\ref{eq: meta-train-t}-\ref{eq: meta-test-t}. 
The third one is to use $\mathcal{L}_{\text{S}}$ and $\mathcal{L}_{\text{W}}$ in a multi-task learning (MTL) manner. 
The first and third methods can get 0.34 and 0.48 gains, respectively. It is unexpected that the MAML does not work. 
We believe one of the reasons is that tasks in two stages of MAML are too similar.

On the other hand, when ablating $\mathcal{L}_{\text{S}}$ and $\mathcal{L}_{\text{W}}$, the BLEU score drops 0.37 and 0.52, respectively. We further attempts to add the $\mathcal{L}_{\text{T}}$ to meta-train step, which leads to a 0.27 drop. We think the reason for this is the same as  MAML. In summary, the CAML algorithm is conducive to training, making the model achieve better results. $\mathcal{L}_{\text{S}}$ and $\mathcal{L}_{\text{W}}$ play different roles and brings certain improvements to the model.

\section{Related Work}
\paragraph{Reliability vs. Robustness.}
There is some overlap between robustness and reliability in NMT~\cite{Sperber2017TowardRN}. Thus, some studies on the robustness of NMT can partially improve reliability. For example, \citet{liu2019robust} proposed to incorporate noisy sentences into NMT training to address \textit{homophone noise}, where words are replaced by others with similar pronunciations. \citet{ebrahimi2018adversarial,cheng2020advaug} proposed to use adversarial examples to address the \textit{synonym problem}. However, robustness focuses on noisy inputs, which can deal with the errors in the sentence. 
Some pre-processing methods, \textit{e.g.}, grammar correction models, are another way to solve the robustness problem. Our work improves the reliability of NMT by addressing the source diversity phenomenon. Various expressions, all of which are proper, are more common and intractable than incorrect inputs.  

\paragraph{Meta-learning in NLP.}
Meta-learning offers a strong generalization ability with limited data. Previous work has successfully employed model-agnostic meta-learning (MAML)~\cite{finn2017model} to various tasks, e.g., text classification~\cite{bansal2020self} and generation~\cite{qian2019domain}. In NMT, \citet{gu2018meta} proposed to improve the low-resource machine translation with MAML. \citet{li2018learning} proposed to improve the domain generalization of NMT with MAML.
The proposed CAML is a derivative of the standard MAML, which focuses on a more fundamental issue of how to solve the problem of source diversity to improve the reliability of NMT.

\paragraph{Self-supervised training.} 
Recently, self-supervised training has been widely used in NLP~\cite{devlin2018bert,brown2020language,lample2019cross}. In NMT, \citet{sennrich2016improving} firstly proposed back-translation to translate target monolingual data to synthetic parallel data. 
\citet{caswell2019tagged} proposed to add a tag at the source side of the synthetic data. These data-augmentation methods objectively enhance the ability of NMT to deal with source diversity. 
\citet{wei2020uncertainty,wei2022learning} proposed to model continuous semantics towards reducing the source uncertainty problem. Following these studies, our work adopts the self-supervised method to learn the consistent representation from semantically equivalent sentences.

\section{Conclusion}
In this work, we clarify the shortcoming of current NMT models, specifically models trained by the one-to-one paradigm are hard to deal with the multi-to-one scenario.
To address this problem, we present a consistency-aware meta-learning (CAML) framework derived from the model-agnostic meta-learning algorithm. In the proposed CAML, two customized training objectives are used with the MAML manner to model the consistent representation, thereby enhancing the reliability of the NMT model.
Extensive experiments show that CAML effectively improves translation quality by producing stable outputs for diverse inputs.

\section*{Acknowledgments}
We would like to thank the reviewers for their insightful comments. This work is supported by the National Natural Science Foundation of China (Grant No. 62036004).

\bibliography{aaai23}

\end{document}